\documentclass[letterpaper]{article} % DO NOT CHANGE THIS
\usepackage{aaai25}  % DO NOT CHANGE THIS
\usepackage{times}  % DO NOT CHANGE THIS
\usepackage{helvet}  % DO NOT CHANGE THIS
\usepackage{courier}  % DO NOT CHANGE THIS
\usepackage[hyphens]{url}  % DO NOT CHANGE THIS
\usepackage{graphicx} % DO NOT CHANGE THIS
\usepackage{natbib}  % DO NOT CHANGE THIS AND DO NOT ADD ANY OPTIONS TO IT
\usepackage{caption} % DO NOT CHANGE THIS AND DO NOT ADD ANY OPTIONS TO IT

\usepackage{booktabs}
\usepackage{subcaption}
\usepackage{amsmath}
\usepackage{amssymb}
\usepackage{amsmath,amssymb,amsfonts}
\usepackage{booktabs}
\usepackage{bm}
\usepackage{multirow}
\usepackage{amssymb}
\usepackage{mathtools}
\usepackage{makecell}
\usepackage{enumitem}
\usepackage{bbding}

\usepackage{algorithm}
\usepackage{algorithmic}

\usepackage{newfloat}
\usepackage{listings}
\DeclareCaptionStyle{ruled}{labelfont=normalfont,labelsep=colon,strut=off} % DO NOT CHANGE THIS
\floatstyle{ruled}
\newfloat{listing}{tb}{lst}{}
\floatname{listing}{Listing}
\title{Multi-Teacher Knowledge Distillation with Reinforcement Learning for Visual Recognition}
\author{
    Chuanguang Yang\textsuperscript{\rm 1}, Xinqiang Yu\textsuperscript{\rm 1,2}, Han Yang\textsuperscript{\rm 1,2}, Zhulin An\textsuperscript{\rm 1}\thanks{Corresponding author}, Chengqing Yu\textsuperscript{\rm 1,2}, Libo Huang\textsuperscript{\rm 1}, Yongjun Xu\textsuperscript{\rm 1}\\
}
\affiliations{
    \textsuperscript{\rm 1}Institute of Computing Technology, Chinese Academy of Sciences, Beijing, China\\
    \textsuperscript{\rm 2}University of Chinese Academy of Sciences, Beijing, China
    \{yangchuanguang,yuxinqiang21s,yanghan22s,anzhulin,yuchengqing22b,huanglibo,xyj\}@ict.ac.cn
}

\usepackage{bibentry}
\begin{document}
	
	\maketitle
	
	\begin{abstract}
		Multi-teacher Knowledge Distillation (KD) transfers diverse knowledge from a teacher pool to a student network. The core problem of multi-teacher KD is how to balance distillation strengths among various teachers. Most existing methods often develop weighting strategies from an individual perspective of teacher performance or teacher-student gaps, lacking comprehensive information for guidance. This paper proposes Multi-Teacher Knowledge Distillation with Reinforcement Learning (MTKD-RL) to optimize multi-teacher weights. In this framework, we construct both teacher performance and teacher-student gaps as state information to an agent. The agent outputs the teacher weight and can be updated by the return reward from the student. MTKD-RL reinforces the interaction between the student and teacher using an agent in an RL-based decision mechanism, achieving better matching capability with more meaningful weights. Experimental results on visual recognition tasks, including image classification, object detection, and semantic segmentation tasks, demonstrate that MTKD-RL achieves state-of-the-art performance compared to the existing multi-teacher KD works.
	\end{abstract}

		\begin{links}
			\link{Code}{https://github.com/winycg/MTKD-RL}
		\end{links}
	\section{Introduction}
	Deep neural networks~\cite{yang2020gated,touvron2021going,liang2024simple,liang2024survey} have achieved excellent performance for visual recognition tasks, but accompanied by the growth of computation complexity and memory footprint. Knowledge Distillation (KD)~\cite{hinton2015distilling,zhang2023generalization} becomes an effective way to improve a low-complexity student network, and has been widely applied to image classification~\cite{yang2022mixskd,yang2023online}, object detection~\cite{lidetkds}, segmentation~\cite{yang2022cross} and generation~\cite{feng2024relational}. The conventional KD often investigates a single teacher to transfer knowledge to a student. Compared to the single-teacher KD, multi-teacher KD~\cite{zhang2022confidence,zhang2023adaptive} provides more comprehensive and diverse knowledge, increasing the upper bound of student performance. 
	
	Although multi-teacher KD could lead to more significant improvements, it is more challenging since balancing distillation strengths among various teachers is a non-trivial problem. The vanilla multi-teacher KD assigns equal weights to distill a student~\cite{you2017learning}. However, the equal distillation strength ignores two critical aspects: (1) \emph{teacher performance}: various teachers may perform differently on the same data sample; (2) \emph{teacher-student gaps}: a more powerful teacher may not result in a better student, because a simple student may not have enough capacity to mimic a large teacher, as discussed by Cho \emph{et al.}~\cite{cho2019efficacy}. Previous multi-teacher KD methods often explore weight generation from an individual perspective of the aspect (1) \emph{teacher performance} (\emph{e.g.} information entropy~\cite{kwon2020adaptive,zhang2022confidence} and logits~\cite{zhang2023adaptive}) or the aspect (2) \emph{teacher-student gaps}, reflecting in gradient space~\cite{du2020agree} and similarity matrix~\cite{liu2020adaptive}. However, these methods only consider a single aspect to generate multi-teacher weights but neglect a comprehensive interaction among the teacher, student, and data samples. 
	\begin{figure*}[tbp]  
		\centering 
		\includegraphics[width=0.85\linewidth]{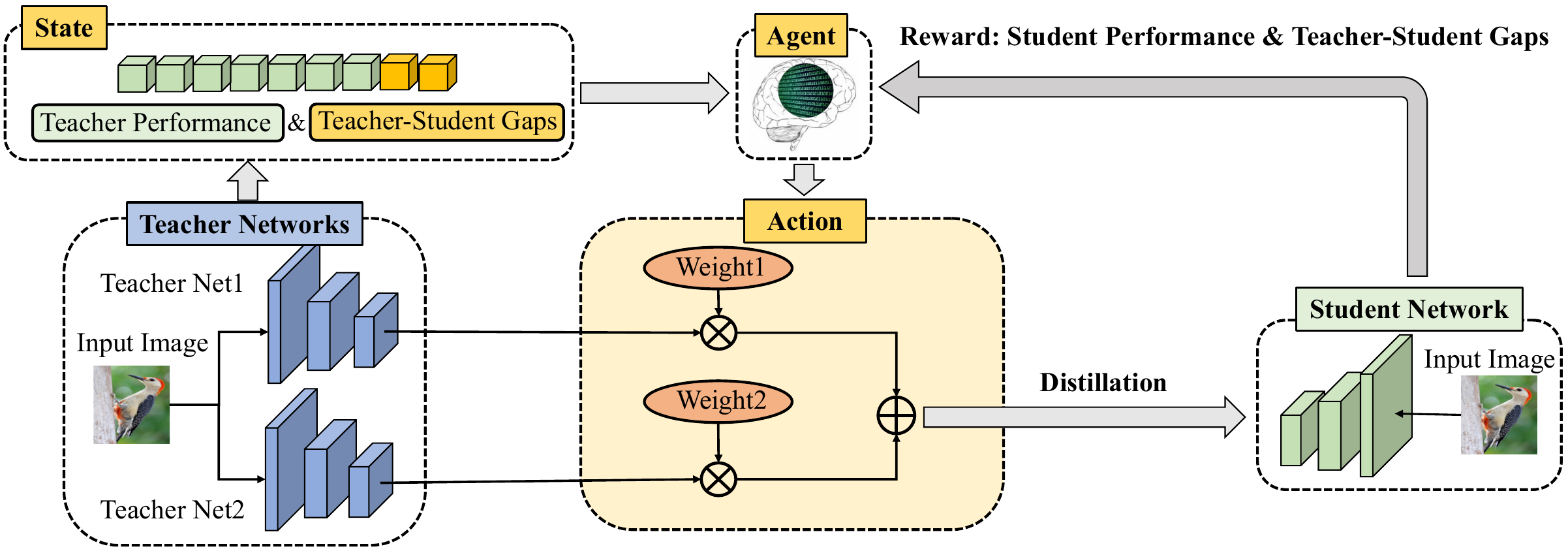}
		\caption{Overview of the basic idea about our proposed MTKD-RL.} 
		\label{overview_self_kd}
	\end{figure*}
	
	This paper proposes Multi-Teacher Knowledge Distillation with Reinforcement Learning (MTKD-RL) to optimize the multi-teacher weight generation problem. The overview of MTKD-RL is illustrated in Fig.\ref{overview_self_kd}. We formulate multi-teacher KD training as a Reinforcement Learning (RL) optimization process. We introduce an agent to output the action of multi-teacher weights according to the state information. The state, including teacher information (feature, logit, and cross-entropy loss) and teacher-student gaps (feature similarity and logit KL-divergence), encodes evidence of  both the aspect (1) and (2), respectively. We use the student performance and teacher-student gaps as rewards to update the agent by policy gradient algorithm. The framework alternatively performs multi-teacher KD and agent optimization until the student is converged. 
	
	Compared to previous multi-teacher KD works, our method constructs teacher performance and teacher-student gaps as prompts, covering the aspect (1) and (2). MTKD-RL reinforces the interaction between the student and teacher using an agent in an RL-based decision mechanism, achieving better matching capability with more meaningful weights. Experimental results on image classification tasks demonstrate that MTKD-RL performs better than previous state-of-the-art multi-teacher KD works. Extensive experiments on downstream object detection and semantic segmentation show that MTKD-RL guides the student to learn better features for dense prediction tasks.

	The contributions are two-fold:
	(1) we propose MTKD-RL, an RL-based method to generate meaningful multi-teacher KD weights by reinforcing the interaction between the student and teacher. (2) Experimental results on visual recognition tasks show that MTKD-RL achieves state-of-the-art performance.

	\section{Related Work}
	\subsection{Multi-Teacher Knowledge Distillation}
	The conventional KD framework~\cite{hinton2015distilling,yang2021hierarchical,yang2022knowledge,yang2022cross,yang2024clip} often improves a student network taught by a high-capacity teacher. However, the student performance may be bounded by a single teacher due to limited knowledge. Many works attempted to rely on multiple teacher networks for knowledge transfer. The core question of multi-teacher KD is how to balance the weights among various teacher networks. A straightforward method~\cite{you2017learning} is to assign an equal weight to each teacher, but it may ignore the diversity of teachers and lead to suboptimum performance. 
	
	Some subsequent works explore better multi-teacher KD algorithms with meaningful balancing weights.  AE-KD~\cite{du2020agree} formulates multi-teacher KD as a multi-objective optimization problem and derives the weight solutions from the perspective of gradient space. AMTML-KD~\cite{liu2020adaptive} introduces a latent factor for each teacher and calculates teacher importance by making an element-wise product with the student max-pooled features. EB-KD~\cite{kwon2020adaptive}  computes aggregation weights according to information-entropy, since the entropy measures the confidence of a teacher predictive distribution. However,  EB-KD may assign an inaccurate weight when the teacher's predictive category is not correct. To address this problem, CA-MKD~\cite{zhang2022confidence} introduces cross-entropy complemented by ground-truth labels to guarantee the correct category guidance. Instead of exploiting probability-level information, the recent MMKD~\cite{zhang2023adaptive} introduces a meta-learning mechanism~\cite{finn2017model} with a hard buffer to optimize aggregation weights of both the teachers' features and logits. Although both MOBA~\cite{kang2022moba} and RMTS~\cite{yuan2021reinforced} involve RL and distillation, they have essential distinctions compared to MTKD-RL in \emph{solved tasks}. MOBA boosts RL models in continuous control tasks, \emph{e.g.} games and robotics. RMTS enhances BERT models for natural language processing tasks. Our MTKD-RL improves CNN or ViT models for visual recognition tasks. More references about multi-teacher KD works could refer to this survey~\cite{yang2023categories}. 
	
	Although previous multi-teacher KD methods achieve good performance gains, they often consider a single level to optimize teacher weights.  Moreover, they neglect the interaction between ensemble teachers and the student, especially student accuracy as a critical indicator. Therefore, the generated weights may not reflect the comprehensive capability of teachers and lack compatibility with a specific student. By contrast, our MTKD-RL method  constructs both the teacher performance and teacher-student gaps as the state, optimized by the student performance as rewards, leading to more meaningful teacher weights.

	\subsection{Reinforcement Learning}
	Reinforcement Learning (RL) has achieved great success in decision-making~\cite{kaelbling1996reinforcement,liu2020new,chengqing2023multi}. The basic framework of the RL system~\cite{mnih2013playing,van2016deep} is to maximize the reward achieved by an agent when interacting with the environment. During RL training, the agent outputs an action using the observed state from the environment.  After action execution, the RL system updates the agent according to the returned reward. The RL loop iterates on episodes until the agent converges.
	
	The RL optimization can be divided into value-based (\emph{e.g. }DQN~\cite{mnih2013playing} and DDQN~\cite{van2016deep}) and policy-based (\emph{e.g. } PG~\cite{sutton1999policy} and PPO~\cite{schulman2017proximal}) algorithms. Value-based RL often selects an action with the maximum Q-value.  Policy-based RL constructs a probability distribution to sample an action. Compared with value-based RL, policy-based RL has two advantages: (1) it is more compatible with continuous action space; (2) it avoids policy degradation since it does not have value function errors in value-based RL. Based on the above analyses, policy-based RL is more suitable for optimizing continuous multi-teacher weights as action space. Policy Gradient (PG)~\cite{sutton1999policy} uses reward from the observed state to optimize the action policy. In the conventional RL system, the original PG may be difficult to converge due to the large gradient variance. Some advanced variants are proposed to improve PG based on actor-critic (\emph{e.g. }DPG~\cite{silver2014deterministic} and DDPG~\cite{lillicrap2015continuous}) and trust region (\emph{e.g. }PPO~\cite{schulman2017proximal}). 
	
	As shown in Table~\ref{ablation}, we found that the multi-teacher decision is not sensitive to PG algorithms. One possible reason is that the gradient optimization of multi-teacher decisions with a fixed visual dataset is more stable than the conventional RL environment. Therefore, we choose the original PG to optimize multi-teacher weights. Notice that this paper focuses on a unified framework of multi-teacher KD with RL instead of proposing a new RL algorithm.
	
	\section{Methodology}
	
	\subsection{Preliminary of Multi-Teacher KD}
	The standard KD transfers knowledge from a teacher network $T$ to a student network $S$. The framework is often formulated as Equ.(\ref{basic_kd}), including a basic task loss, logit KD loss, and feature KD loss.
	\begin{equation}
		\mathcal{L}_{KD}=\underbrace{\mathcal{H}(\bm{y}_{i}^{S},\bm{y}_{i})}_{task\ loss}+\alpha \underbrace{\mathcal{D}_{KL}(\bm{y}_{i}^{S},\bm{y}_{i}^{T})}_{logit\ KD\ loss}+\beta \underbrace{\mathcal{D}_{dis}(\bm{F}_{i}^{S},\bm{F}_{i}^{T})}_{feature\ KD\ loss}.
		\label{basic_kd}
	\end{equation}
	$\bm{y}_{i}$ is the ground-truth label of the $i$-th sample. $\bm{y}_{i}^{S}$ and $\bm{y}_{i}^{T}$ are predictive logits from the student and teacher, respectively. $\bm{F}_{i}^{S}$ and $\bm{F}_{i}^{T}$ are features from the student and teacher,  respectively. $\mathcal{H}$ denotes cross-entropy function. $\mathcal{D}_{KL}$ indicates Kullback-Leibler divergence to measure the discrepancy of student and teacher probability distributions. $\mathcal{D}_{dis}$ is a distance metric to measure the similarity of student and teacher features. $\alpha$ and $\beta$ are loss weights for logit KD and feature KD losses, respectively.
	
	When extending to multi-teacher KD, each teacher network transfers knowledge to the same student network, resulting in the total loss as Equ.(\ref{mtkd_basic_kd}).
	\begin{align}
		\mathcal{L}_{MTKD}=&\underbrace{\mathcal{H}(\bm{y}_{i}^{S},\bm{y}_{i})}_{task\ loss}+\alpha \underbrace{\sum_{m=1}^{M}w_{l,i}^{m}\mathcal{D}_{KL}(\bm{y}_{i}^{S},\bm{y}_{i}^{T_{m}})}_{logit\ KD\ loss} \notag \\
		&+\beta \underbrace{\sum_{m=1}^{M}w_{f,i}^{m}\mathcal{D}_{dis}(\bm{F}_{i}^{S},\bm{F}_{i}^{T_{m}})}_{feature\ KD\ loss}.
		\label{mtkd_basic_kd}
	\end{align}
	Here, $M$ is the number of networks in the teacher pool $\{T_{m}\}_{m=1}^{M}$.  This paper focuses on optimizing the teacher weights $\{\bm{w}_{i}^{m}=[w_{l,i}^{m},w_{f,i}^{m}]\}_{m=1}^{M}$ using reinforcement learning, as shown in the next section. Without bells and whistles, we use the traditional KD~\cite{hinton2015distilling} as the logit KD loss and FitNet~\cite{romero2014fitnets} as the feature KD loss by default. We set $\alpha=1$ and $\beta=5$, as analysed in Fig.\ref{Gate}. More advanced logit KD losses (\emph{e.g.} DIST~\cite{huang2022knowledge}) and feature KD losses (\emph{e.g.} ND~\cite{wang2025improving}) could also be applied in Equ.(\ref{mtkd_basic_kd}), which are orthogonal to our multi-teacher KD framework.

	\subsection{RL-based Multi-Teacher  Weights Optimization}
	In this section, we propose to optimize sample-wise teacher weights dynamically using reinforcement learning. As shown in Fig.\ref{overview_self_kd}, we introduce an agent over the training graph. The agent interacts with the multi-teacher KD environment. It receives the state information to generate teacher weights as the action. We apply teacher weights into Equ.(\ref{mtkd_basic_kd}) to distill the student network. After an episode of training samples, the student performance is regarded as the reward to optimize agent parameters. The reinforced loop proceeds until the student network converges.  The instantiation of (\textbf{state},\textbf{action},\textbf{reward}) in the context of multi-teacher KD is shown as follows.
	
	\subsubsection{State} In our RL-based framework, the state representation of the input sample $\bm{x}_{i}$ is formulated as an embedding $\bm{s}_{i}$. The state $\bm{s}_{i}$ is a concatenation of teacher performance (\emph{i.e.} feature representation, logit  vector, and cross-entropy loss) and  teacher-student gaps (\emph{i.e.} feature similarity and probability KL-divergence). The concrete details are formulated as follows.
	
	\textbf{(1) Teacher feature representation.} Given an input sample $\bm{x}_{i}$, the $m$-th teacher's feature representation after the penultimate layer is defined as $\bm{f}_{i}^{T_m}\in \mathbb{R}^{d_m}$, where $d_m$ is the embedding size of the $m$-th teacher. The feature representation often encodes semantic information from the input sample. 
	
	\textbf{(2) Teacher logit  vector.} Given an input sample $\bm{x}_{i}$, the $m$-th teacher's logit vector after the final output layer is defined as $\bm{z}_{i}^{T_m}\in \mathbb{R}^{C}$, where $C$ is the number of classes. The logit vector represents the direct prediction of a class confidence distribution over the complete class space.
	
	\textbf{(3) Teacher cross-entropy loss.} Given an input sample $\bm{x}_{i}$ with the label $\bm{y}_{i}$, the $m$-th teacher's cross-entropy loss is defined as $\mathcal{L}_{CE}^{T_{m}}=\mathcal{H}(\bm{y}_{i}^{T_{m}},\bm{y}_{i})$. The cross-entropy loss $\mathcal{L}_{CE}^{T_{m}}$ measures the fitting performance between the teacher's prediction and the ground-truth label.
	
	\textbf{(4) Teacher-student feature similarity.}  Given an input sample $\bm{x}_{i}$, the cosine similarity between the student's feature representation $\bm{f}_{i}^{S}\in \mathbb{R}^{d}$  and the teacher's feature representation $\bm{f}_{i}^{T_m}\in \mathbb{R}^{d_m}$ is formulated as cosine similarity:
	\begin{equation}
		\cos_{i}^{T_m}= r(\bm{f}_{i}^{S})\cdot \bm{f}_{i}^{T_m}/  || r(\bm{f}_{i}^{S})|| || \bm{f}_{i}^{T_m} ||,
		\label{feat_sim}
	\end{equation}
	where $r(\cdot)$ is a linear regressor to transform $\bm{f}_{i}^{S}$ to match the embedding size of $\bm{f}_{i}^{T_m}$. $\cdot$ is a dot product and $||\cdot||$ denotes $l_2$ norm. The  cosine similarity measures the distance gap between the student and teacher in feature space.
	
	\textbf{(5) Teacher-student probability KL-divergence.}  Given an input sample $\bm{x}_{i}$, the KL-divergence between the student's logit vector $\bm{y}_{i}^{S}\in \mathbb{R}^{C}$  and the teacher's logit vector $\bm{y}_{i}^{T_m}\in \mathbb{R}^{C}$ is formulated as:
	\begin{equation}
		KL_{i}^{T_m}=\mathcal{D}_{KL}(\bm{y}_{i}^{S},\bm{y}_{i}^{T_{m}}).
		\label{t_s_logit_div}
	\end{equation}
	 The KL-divergence measures  the discrepancy between the student and teacher in class probability space.
	
	We concatenate (1)$\sim$(5) to construct the state embedding. Given an input sample $\bm{x}_{i}$, the $m$-th teacher's state embedding is formulated as $\bm{s}_{i}^{m}$:
	\begin{equation}
		\bm{s}_{i}^{m}=[\underbrace{\bm{f}_{i}^{T_m}\parallel  \bm{z}_{i}^{T_m}\parallel  \mathcal{L}_{CE}^{T_{m}}}_{Teacher\ Performance}\parallel \underbrace{\cos_{i}^{T_m}\parallel KL_{i}^{T_m}}_{Teacher-Student\ Gaps}],
		\label{state_info}
	\end{equation}
	where $\parallel $ denotes the concatenation operator. Based on comprehensive state information, the agent can make dynamic decisions by referring to teacher performance and teacher-student gaps among various input samples.
	
	\subsubsection{Action} We construct an agent $\pi_{\bm{\theta}_m}(\bm{s}_{i}^{m})$ for each teacher network $T_{m}$, where $\bm{\theta}_m$ denotes the trainable agent parameters. The agent model $\pi_{\bm{\theta}_m}(\bm{s}_{i}^{m})$ contains several linear layers with a middle ReLU activation function, and finished by a softmax function. The output is the teacher weight vector $\bm{w}_{i}^{m}$, each weight is within $(0,1)$  in continuous action space. The process is formulated as $\bm{w}_{i}^{m}=\pi_{\bm{\theta}_m}(\bm{s}_{i}^{m})$. We also combine the  confidence- and divergence-aware weight generation strategies in the action space. The detailed action construction is shown in appendix.

	\subsubsection{Reward} 
	We define an episode as one training batch $\{\bm{x}_i\}_{i=1}^{B}$ including $B$ samples. For each sample  $\bm{x}_i$, we construct the state embedding $\bm{s}_{i}^{m}$ for each teacher network $T_{m}$ and  guide the agent $\pi_{\bm{\theta}_m}(\bm{s}_{i}^{m})$ to produce the teacher weight $\bm{w}_{i}^{m}$. We perform weighted multi-teacher KD (Equ.(\ref{mtkd_basic_kd})) with the generated weights $\{\bm{w}_{i}^{m}\}_{m=1}^{M}$ to train the student network. The reward function should be relevant to the trained student performance after multi-teacher KD weighted by the agent. A better student network should have a lower classification loss and more similar class probability and feature distributions with the teacher. Therefore, we utilize cross-entropy loss between the student and ground-truth labels, probability KL-divergence loss and feature Mean Squared Error (MSE) loss between the student and teacher as evidence to construct the reward function as Equ.(\ref{reward}):
	\begin{equation}
		R_{i}^{m}=-\mathcal{H}(\bm{y}_{i}^{S},\bm{y}_{i})-\alpha\mathcal{D}_{KL}(\bm{y}_{i}^{S},\bm{y}_{i}^{T_{m}})-\beta \mathcal{D}_{dis}(\bm{F}_{i}^{S},\bm{F}_{i}^{T_{m}}),
		\label{reward}
	\end{equation}
	where $R_{i}^{m}$ denotes the reward of the $m$-th agent over the $i$-th input sample. Here, the reward is defined by the negative loss, since a lower loss value indicates a better student.
	
	\subsubsection{Optimize agent with policy gradient}  
	Unlike traditional supervised learning, RL-based optimization often does not know whether the chosen action is correct or not. Therefore, policy-based RL often optimizes the agent according to the returned reward after making a decision. If an agent earns a larger reward after choosing an action, Policy Gradient (PG)~\cite{sutton1999policy} would increase the corresponding gradient derived from this action. We apply PG to optimize the agent's parameters, as shown in Equ.(\ref{gradient_update}).
	\begin{equation}
		\bm{\theta}_m \longleftarrow \bm{\theta}_{m}-\eta\sum_{i=1}^{B}\bar{R}_{i}^{m}\bigtriangledown_{ \bm{\theta}_{m}}\pi_{\bm{\theta}_m}(\bm{s}_{i}^{m}),\ m=1,2,\cdots,M.
		\label{gradient_update}
	\end{equation}
	$\bar{R}_{i}^{m}$ is the normalized reward formulated as Equ.(\ref{reward_norm}). 
	\begin{equation}
		\bar{R}_{i}^{m}=\frac{R_{i}^{m}-\min_{k}R_{i}^{k}}{\max_{k}R_{i}^{k}-\min_{k}R_{i}^{k}}-\frac{1}{M}\sum_{k=1}^{M}R_{i}^{k}.
		\label{reward_norm}
	\end{equation}
	Here, $k\in \{1,\cdots,M\}$. We apply min-max normalization to rescale the reward $R_{i}^{m}$ to $(0,1)$, and then subtract the mean value of $M$ teacher rewards $\{R_{i}^{k}\}_{k=1}^{M}$.  The PG would lead to a positive update when $\bar{R}_{i}^{m}>0$, or a negative update when $\bar{R}_{i}^{m}<0$. Benefiting from PG optimization, the agent prefers adaptively assigning teacher weights according to the reward.
	
	\begin{algorithm}[t]
		\small
		1. Pre-train the student network $S$ by multi-teacher KD following $\mathcal{L}_{MTKD}$ (Equ.(\ref{mtkd_basic_kd})) with equal weights, \emph{i.e.} $\{\bm{w}_i^{m}=\bm{1}\}_{m=1}^{M}$, using all training samples $\bm{x}_{i}\in \mathcal{D}$.
		
		2. Pre-train the agent models $\{\pi_{\bm{\theta}_m}\}_{m=1}^{M}$ using the returned rewards $\{R_{i}^{m}\}_{m=1}^{M}$  with $\{\bm{w}_{i}^{m}=\bm{1}\}_{m=1}^{M}$ as the action for all training samples $\bm{x}_{i}\in \mathcal{D}$.
		
		3. Run Algorithm~\ref{alg:rl-ts} to alternatively perform multi-teacher KD and agent optimization until convergence.
		\caption{Overall MTKD-RL Training Procedure}
		\label{alg:complete}
	\end{algorithm}

	\begin{algorithm}[t]
		\caption{Alternative Multi-Teacher KD and Agent Optimization}
		\label{alg:rl-ts}
		\textbf{Input}: Training dataset $\mathcal{D}$. Pre-trained teacher networks $\{T_{m}\}_{m=1}^{M}$.  A student network $S$.  Agent models $\{\pi_{\bm{\theta}_m}\}_{m=1}^{M}$. \\
		\textbf{Output}: Trained student $S$. 
		\begin{algorithmic}[1] 
			\WHILE{the student network $S$ is not converged}   
			\STATE Randomly Shuffle $\mathcal{D}$ to produce a new batch sequence. 
			\FOR{each batch $\mathcal{B} \in \mathcal{D}$} 
			\STATE Construct state information $\{\bm{s}_{i}^{m}\}_{m=1}^{M}$ following Equ.(\ref{state_info}) for each sample $\bm{x}_i \in \mathcal{B}$, where the batch size is $B$. 
			\STATE Generate multi-teacher weights $\{\bm{w}_{i}^{m}\}_{m=1}^{M}$ by $\bm{w}_{i}^{m}=\pi_{\bm{\theta}_m}(\bm{s}_{i}^{m})$ 
			\STATE Compute the multi-teacher KD loss $\mathcal{L}_{MTKD}$ following Equ.(\ref{mtkd_basic_kd}) with generated weights $\{\bm{w}_{i}^{m}\}_{m=1}^{M}$. 
			\STATE Update the student network $S$ using $\mathcal{L}_{MTKD}$. 
			\STATE Compute rewards $\{R_{i}^{m}\}_{m=1}^{M}$ following Equ.(\ref{reward}).
			\STATE Save $(\{\{\bm{s}_{i}^{m}\}_{m=1}^{M},\{\bm{w}_{i}^{m}\}_{m=1}^{M},\{R_{i}^{m}\}_{m=1}^{M}\}_{i=1}^{B})$ to the episode history $\mathcal{H}$.
			\ENDFOR
			\FOR{each $(\{\{\bm{s}_{i}^{m}\}_{m=1}^{M},\{\bm{w}_{i}^{m}\}_{m=1}^{M},\{R_{i}^{m}\}_{m=1}^{M}\}_{i=1}^{B}) \in \mathcal{H}$}
			\STATE Update the agent models $\{\pi_{\bm{\theta}_m}\}_{m=1}^{M}$ following Equ.(\ref{gradient_update}).
			\ENDFOR
			\ENDWHILE
		\end{algorithmic}
	\end{algorithm}

	\subsection{Overall MTKD-RL Framework}
	Algorithm~\ref{alg:complete} illustrates the overall MTKD-RL training procedure. At first, we pre-train the student network $S$ by multi-teacher KD following $\mathcal{L}_{MTKD}$ (Equ.(\ref{mtkd_basic_kd})) with equal weights for one training epoch. Afterwards, we use the collected (state, action, reward) information during multi-teacher KD to pre-train the agent models $\{\pi_{\bm{\theta}_m}\}_{m=1}^{M}$. After pre-training, the RL-loop would start from a good initialization and avoid the training collapse problem. Then we run Algorithm~\ref{alg:rl-ts} to  alternatively perform multi-teacher KD and agent optimization until convergence.
	
	As shown in Algorithm~\ref{alg:rl-ts}, for each iterative epoch, we freeze the agent models $\{\pi_{\bm{\theta}_m}\}_{m=1}^{M}$ and train the student network $S$ using multi-teacher KD with the generated weights from agent models. Then, we utilize the collected (state, action, reward) information during multi-teacher KD to optimize the agent models $\{\pi_{\bm{\theta}_m}\}_{m=1}^{M}$. The alternative training proceeds until the student network $S$ is converged.

	\section{Experiments}
	\subsection{Experimental Setup}
	\textbf{Dataset.} We use CIFAR-100~\cite{krizhevsky2009learning} and ImageNet~\cite{deng2009imagenet} datasets for image classification experiments.   The object detection experiments adopt COCO-2017~\cite{lin2014microsoft} dataset. We utilize Cityscapes~\cite{cordts2016cityscapes}, ADE20K~\cite{zhou2017scene} and COCO-Stuff-164K~\cite{caesar2018coco} datasets for semantic segmentation.
	
	\textbf{Compared methods.} We compare several methods in this experimental section. 'Baseline' denotes the traditional training without distillation. 'AVER (KD+FitNet)' means the multi-teacher KD with equal weights for the traditional Hinton's logit KD loss~\cite{hinton2015distilling}  and Mean Squared Error (MSE)-based feature KD loss~\cite{romero2014fitnets}. Other compared methods, such as  AMTML-KD~\cite{liu2020adaptive}, AEKD~\cite{du2020agree}, CA-MKD~\cite{zhang2022confidence}, and MMKD~\cite{zhang2023adaptive}, are advanced strategies to compute multi-teacher KD weights.
	
	\textbf{Evaluated networks.} Experiments are conducted over Convolutional Neural Networks (CNNs) and Vision Transformers (ViTs).  CNNs includes ResNet~\cite{he2016deep}, WRN~\cite{zagoruyko2016wide}, ResNeXt~\cite{xie2017aggregated}, MobileNetV2~\cite{sandler2018mobilenetv2}, ShuffleNetv2~\cite{ma2018shufflenet}, and RegNet~\cite{radosavovic2020designing}. ViTs includes CaiT~\cite{touvron2021going} and DeiT~\cite{touvron2021training}.
	\emph{More training details are shown in appendix.}
	\subsection{Experimental Results on Image Classification}
	\textbf{Experiments on CIFAR-100.} In Table~\ref{CIFAR_cnn_mkd}, we conduct multi-teacher KD experiments on CIFAR-100. Compared to the baseline without distillation, even the vanilla AVER (KD) with equal weights achieves significant improvements over four student networks with an average gain of 2.46\%. AVER (KD+FitNet) complements feature-level distillation and obtains an average improvement of 0.68\% over the logit-level AVER (KD). Previous multi-teacher weight generation strategies, such as AMTML-KD, AEKD, CA-MKD, and MMKD, generally outperform the AVER version, indicating that dynamic sample-wise weights in a data-driven manner are better than simple static weights. Our MTKD-RL achieves the best performance and surpasses the state-of-the-art CA-MKD and MMKD methods with average gains of 0.31\% and 0.33\%, respectively. The results demonstrate that our proposed reinforcement learning framework is better than the entropy-based and meta-learning-based mechanisms for multi-teacher weight optimization.

	\begin{table}[t!]
		\centering
		
		\resizebox{1\linewidth}{!}{
			\begin{tabular}{l |c c |l|c }
				\toprule
				Network & Params & FLOPs & Method & Acc@1 \\
				
				\midrule 
				\multicolumn{2}{@{} l}{Teacher pool}  \\
				\midrule
				RegNetY-400MF& 	5.8M &467.1M &\multirow{4}{*}{Pretrained}  & 78.87  \\
				RegNetX-400MF&  4.8M & 466.2M & 	&
				79.15  \\
				ResNet-32x4&  7.4M & 1083.0M & 	& 79.59  \\
				WRN-28-4 &5.9M &  845.6M & &	79.37  \\
				\midrule
				\multicolumn{2}{@{} l}{Student}  \\
				\midrule
				\multirow{8}{*}{RegNetX-200MF} & \multirow{8}{*}{2.4M} & \multirow{8}{*}{223.4M}  & Baseline & 77.38$_{\pm 0.23}$  \\
				&&& AVER (KD) & 78.64$_{\pm 0.31}$ \\
				&&& AVER (KD+FitNet) & 79.12$_{\pm 0.28}$\\
				&&& AMTML-KD& 79.46$_{\pm 0.45}$ \\
				&&& AEKD & 79.22$_{\pm 0.37}$\\
				&&& CA-MKD& \underline{80.28}$_{\pm 0.26}$\\
				&&& MMKD& 80.15$_{\pm 0.41}$ \\
				&&& MTKD-RL  (Ours) & \textbf{80.58}$_{\pm 0.29}$  \\
				
				\midrule
				\multirow{8}{*}{MobileNetV2} & \multirow{8}{*}{2.4M} & \multirow{8}{*}{22.4M}  & Baseline & 69.17$_{\pm 0.18}$ \\
				&&& AVER (KD) & 71.85$_{\pm 0.27}$ \\
				&&& AVER (KD+FitNet) &72.67$_{\pm 0.26}$ \\
				&&& AMTML-KD & 72.89$_{\pm 0.36}$ \\
				&&& AEKD & 72.82$_{\pm 0.42}$ \\
				&&& CA-MKD&74.16$_{\pm 0.32}$ \\
				&&& MMKD& \underline{74.35}$_{\pm 0.34}$ \\
				&&& MTKD-RL  (Ours) &\textbf{74.63}$_{\pm 0.35}$ \\

				\midrule
				\multirow{8}{*}{ShuffleNetv2} & \multirow{8}{*}{1.4M} & \multirow{8}{*}{44.5M}  & Baseline & 72.84$_{\pm 0.15}$ \\
				&&& AVER (KD) & 75.77$_{\pm 0.18}$ \\
				&&& AVER (KD+FitNet) &76.83$_{\pm 0.34}$ \\
				&&& AMTML-KD & 76.93$_{\pm 0.26}$ \\
				&&& AEKD & 77.08$_{\pm 0.21}$ \\
				&&& CA-MKD&\underline{78.09}$_{\pm 0.17}$ \\
				&&& MMKD& 77.87$_{\pm 0.20}$ \\
				&&& MTKD-RL  (Ours) &\textbf{78.39}$_{\pm 0.14}$ \\
				
				\midrule
				\multirow{8}{*}{ResNet-56} & \multirow{8}{*}{0.9M} & \multirow{8}{*}{125.8M}  & Baseline & 72.52$_{\pm 0.27}$ \\
				&&& AVER (KD) & 73.57$_{\pm 0.29}$ \\
				&&& AVER (KD+FitNet) &73.93$_{\pm 0.16}$\\
				&&& AMTML-KD & 74.21$_{\pm 0.33}$ \\
				&&& AEKD & 74.02$_{\pm 0.39}$ \\
				&&& CA-MKD&75.17$_{\pm 0.24}$ \\
				&&& MMKD& \underline{75.26}$_{\pm 0.27}$ \\
				&&& MTKD-RL  (Ours) &\textbf{75.35}$_{\pm 0.33}$ \\
				
				\bottomrule
		\end{tabular}}
	\caption{\textbf{Comparison of accuracy among various multi-teacher KD methods over CNNs  on CIFAR-100. }We use the pre-trained teacher pool to distill student networks. The \textbf{bold}  and underline numbers denotes the best and second-best results for each student. Experiments perform 3 runs.}
		
		\label{CIFAR_cnn_mkd}
	\end{table}

	\begin{table}[t!]
		\centering
		
		\resizebox{1\linewidth}{!}{
			\begin{tabular}{l |c c |l|c }
				\toprule
				Network & Params & FLOPs & Method & Acc@1 \\
				
				\midrule 
				\multicolumn{2}{@{} l}{Teacher pool}  \\
				\midrule
				ResNet-50& 	25.6M &4.1G &\multirow{4}{*}{Pretrained}  & 76.13  \\
				ResNet-101&  44.5M& 7.8G & 	&
				77.37  \\
				Wide ResNet-50-2&  68.9M & 11.4G & 	&78.47  \\
				ResNeXt-50 (32×4d)&25.0M &  4.2G & &	77.62 \\
				\midrule
				\multicolumn{2}{@{} l}{Student}  \\
				\midrule
				\multirow{8}{*}{ResNet-18} & \multirow{8}{*}{11.7M} & \multirow{8}{*}{1.8G}  & Baseline & 70.35  \\
				&&& AVER (KD) & 71.52 \\
				&&& AVER (KD+FitNet) & 71.56\\
				&&& AMTML-KD& 71.89\\
				&&& AEKD & 71.67\\
				&&& CA-MKD& \underline{72.38}\\
				&&& MMKD& 72.33\\
				&&& MTKD-RL  (Ours) &  \textbf{72.82} \\
				
				\midrule
				\multirow{8}{*}{ResNet-34} & \multirow{8}{*}{21.8M} & \multirow{8}{*}{3.7G}  & Baseline & 73.64 \\
				&&& AVER (KD) & 75.32\\
				&&& AVER (KD+FitNet) &75.55\\
				&&& AMTML-KD & 75.68 \\
				&&& AEKD & 75.66\\
				&&& CA-MKD&75.87 \\
				&&& MMKD& \underline{76.06}\\
				&&& MTKD-RL  (Ours) &\textbf{76.77} \\

				\bottomrule
		\end{tabular}}
		\caption{Comparison of accuracy among various multi-teacher KD methods over CNNs  on ImageNet. }
		\label{imagenet_cnn_mkd}
	\end{table}

	\textbf{Experiments on ImageNet.} 
	As shown in Table~\ref{imagenet_cnn_mkd}, we conduct multi-teacher KD experiments over CNNs on ImageNet. MTKD-RL enhances the baseline by 2.47\% and 3.13\% accuracy gains on ResNet-18 and ResNet-34, respectively. It outperforms the AVER (KD+FitNet) with an average gain of 1.24\%, manifesting that our method generates meaningful multi-teacher weights on large-scale ImageNet. Compared to state-of-the-art MMKD, MTKD-RL achieves 0.49\% and 0.71\% accuracy improvements on ResNet-18 and ResNet-34, respectively. The results show the superiority of MTKD-RL for applying to the large-scale dataset compared to other multi-teacher KD strategies. As shown in Table~\ref{imagenet_cnn_mkd}, we further conduct multi-teacher KD experiments over ViTs. MTKD-RL increases the baseline by 2.91\% and 1.90\% on DeiT-Tiny and CaiT-XXS24, respectively. It also exceeds the best competitor MMKD by an average gain of 0.80\% on the two networks. The results reveal that MTKD-RL can also work well on ViT.

	\begin{table}[t!]
		\centering
		
		\resizebox{1\linewidth}{!}{
			\begin{tabular}{l |c c |l|c }
				\toprule
				Network & Params & FLOPs & Method & Acc@1 \\
				
				\midrule 
				\multicolumn{2}{@{} l}{Teacher pool}  \\
				\midrule
				CaiT-S24& 	46.9M &9.4G &\multirow{3}{*}{Pretrained}  & 83.36  \\
				DeiT-Small&  22.1M& 4.6G & 	&
				79.82  \\
				DeiT-Base &  86.6M & 17.5G& 	& 81.80  \\
				\midrule
				\multicolumn{2}{@{} l}{Student}  \\
				\midrule
				\multirow{8}{*}{DeiT-Tiny }& \multirow{8}{*}{5.7M} & \multirow{8}{*}{1.3G}  & Baseline & 72.23  \\
				&&& AVER (KD) & 73.87\\
				&&& AVER (KD+FitNet)  & 73.98 \\
				&&& AMTML-KD& 73.68\\
				&&& AEKD& 73.72\\
				&&& CA-MKD&74.12 \\
				&&& MMKD&\underline{74.35} \\
				&&& MTKD-RL  (Ours) & \textbf{75.14} \\
				\midrule
				\multirow{8}{*}{CaiT-XXS24} & \multirow{8}{*}{12.0M} & \multirow{8}{*}{2.5G}  & Baseline & 77.32 \\
				&&& AVER & 78.36\\
				&&& AVER (KD+FitNet) &78.27 \\
				&&& AMTML-KD &78.57 \\
				&&& AEKD &78.44 \\
				&&& CA-MKD& \underline{78.65}\\
				&&& MMKD& 78.42\\
				&&& MTKD-RL  (Ours) & \textbf{79.22}  \\

				\bottomrule
		\end{tabular}}
		\caption{Comparison of accuracy among various multi-teacher KD methods over ViT  on ImageNet. }
		\label{vit_cnn_mkd}
	\end{table}

	\subsection{Experimental Results on Object Detection} 
	As shown in Table~\ref{mtkd_det}, we transfer ImageNet-pretrained ResNet backbones for downstream object detection on COCO-2017. The ResNet backbones pretrained by our MTKD-RL obtain consistent performance improvements over various detectors (Mask-RCNN~\cite{he2017mask}, Cascade-RCNN~\cite{cai2018cascade}, RetinaNet~\cite{lin2017focal} and Faster-RCNN~\cite{ren2016faster}) than the baseline backbones. MTKD-RL outperforms the baseline with  1.1\% and 1.5\% mAP gains on average for ResNet-18 and ResNet-34, respectively. The results indicate that MTKD-RL can guide the network to learn better feature representations for downstream object detection.

	\begin{table}[t!]
		\centering
		
		\resizebox{1\linewidth}{!}{
			\begin{tabular}{c| c | l |c }
				\toprule
				Backbone &  Detector &   Method & mAP\\
				\midrule
				\multirow{8}{*}{ResNet-18} &  \multirow{2}{*}{Mask-RCNN} & Baseline & 34.1   \\
				& & MTKD-RL (Ours) & \textbf{35.1}  \\
				\cmidrule(r){2-4}
				&  \multirow{2}{*}{Cascade-RCNN} & Baseline & 36.5   \\
				& & MTKD-RL (Ours) & \textbf{37.7}   \\
				\cmidrule(r){2-4}
				&  \multirow{2}{*}{RetinaNet} & Baseline & 31.6   \\
				& & MTKD-RL (Ours) & \textbf{32.7}   \\
				\cmidrule(r){2-4}
				&  \multirow{2}{*}{Faster-RCNN} & Baseline & 33.5  \\
				& & MTKD-RL (Ours) & \textbf{34.7}   \\
				\midrule
				\multirow{8}{*}{ResNet-34} &  \multirow{2}{*}{Mask-RCNN} & Baseline & 37.6   \\
				& & MTKD-RL (Ours) & \textbf{39.0}  \\
				\cmidrule(r){2-4}
				&  \multirow{2}{*}{Cascade-RCNN} & Baseline & 39.6   \\
				& & MTKD-RL (Ours) & \textbf{40.8}   \\
				\cmidrule(r){2-4}
				&  \multirow{2}{*}{RetinaNet} & Baseline & 35.2   \\
				& & MTKD-RL (Ours) & \textbf{36.9}   \\
				\cmidrule(r){2-4}
				&  \multirow{2}{*}{Faster-RCNN} & Baseline & 37.0  \\
				& & MTKD-RL (Ours) & \textbf{38.5}   \\
				
				\bottomrule
		\end{tabular}}
		\caption{Comparison of downstream object detection based on ImageNet-pretrained ResNet backbones over various detectors on COCO-2017. }
		\label{mtkd_det}
	\end{table}
	
	\begin{table}[t!]
		\centering
		
		\resizebox{1\linewidth}{!}{
			\begin{tabular}{l |l| c c c }
				\toprule
				Segmentor  & Method & Cityscapes &  ADE20K & COCO-Stuff \\
				\midrule
				\multirow{2}{*}{DeepLabV3} & Baseline &  76.34 & 36.08 & 29.97\\
				& MTKD-RL (Ours) &  \textbf{77.42} &  \textbf{37.07} &  \textbf{31.82} \\
				\midrule
				\multirow{2}{*}{PSPNet} & Baseline & 74.60 & 36.84 & 31.25 \\
				& MTKD-RL (Ours) &  \textbf{75.89} &  \textbf{37.78} &  \textbf{32.39} \\

				\bottomrule
		\end{tabular}}
		\caption{Comparison of downstream semantic segmentation tasks based on an ImageNet-pretrained ResNet-34 on Cityscapes, ADE20K and COCO-Stuff-164K.}
		\label{seg_cnn_mkd}
	\end{table}

		\begin{figure*}[tbp]  
		\centering 
		\includegraphics[width=1\linewidth]{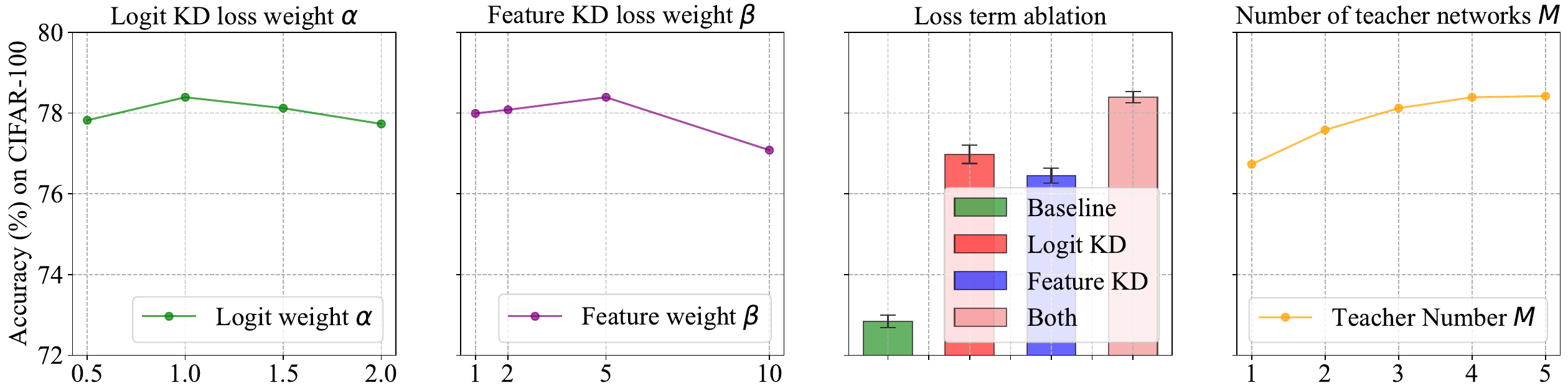}
		\caption{Parameter analyses and ablation study over ShuffleNetV2 on CIFAR-100.} 
		\label{Gate}
	\end{figure*}

		\begin{table*}[t]
		
		\begin{minipage}{0.235\linewidth}
			\centering
			
			\begin{tabular}{l|c}
				\toprule
				RL methods & Acc \\ 
				\midrule 
				PG& \textbf{78.39}$_{\pm 0.14}$  \\
				DPG& 78.16$_{\pm 0.11}$ \\
				DDPG& 78.05$_{\pm 0.22}$\\
				PPO& 78.28$_{\pm 0.08}$ \\
				\bottomrule
			\end{tabular}
			\textbf{(a)} RL methods
			
		\end{minipage}
		\hfill
		\begin{minipage}{0.235\linewidth}
			\centering
			
			\begin{tabular}{l|c}
				\toprule
				Component & Acc \\ 
				\midrule 
				AVER & 76.83$_{\pm 0.34}$  \\
				$\bm{f}_{i}^{T_m}\parallel  \bm{z}_{i}^{T_m}\parallel  \mathcal{L}_{CE}^{T_{m}}$ & 77.84$_{\pm 0.17}$  \\
				$\cos \parallel \mathcal{D}_{KL}$ & 77.46$_{\pm 0.21}$  \\
				All&  \textbf{78.39}$_{\pm 0.14}$ \\
				\bottomrule
			\end{tabular}
			\textbf{(b)} Component analysis
			
		\end{minipage}
		\hfill
		\begin{minipage}{0.4\linewidth}  
			\centering

			\begin{tabular}{c|cc|c}
				\toprule
				Method & Time & Mem & Acc \\ \midrule 
				Baseline & 29s& 2.3G & 72.84$_{\pm 0.15}$ \\
				AVER & 41s &2.8G &  76.83$_{\pm 0.34}$  \\
				CA-MKD & 54s&2.9G &78.09$_{\pm 0.17}$ \\
				MTKD-RL & 47s & 3.2G & \textbf{78.39}$_{\pm 0.14}$ \\
				\bottomrule
			\end{tabular}
			
			\textbf{(c)} Training costs
			
		\end{minipage}
		\caption{Analysis of MTKD-RL over ShuffleNetV2 on CIFAR-100.}
		\label{ablation}
	\end{table*}

	\subsection{Experimental Results on Semantic Segmentation}
	As shown in Table~\ref{seg_cnn_mkd}, we transfer the ImageNet-pretrained ResNet-34 backbone for downstream semantic segmentation datasets. Our MTKD-RL consistently surpasses the baseline across various datasets by equipping with DeepLabV3~\cite{chen2018encoder} and PSPNet~\cite{zhao2017pyramid} segmentation heads.  The average improvements are 1.19\%, 0.97\% and 1.50\% on Cityscapes, ADE20K and COCO-Stuff-164K datasets, respectively. The results show that the backbone pretrained by MTKD-RL can be well extended to semantic segmentation tasks and generate better pixel-wise feature representations.

	\subsection{Ablation Study and Parameter Analysis}
	\textbf{Ablation study of various RL methods.} As shown in Table~\ref{ablation}(a), we investigate various RL methods for multi-teacher weight optimization. We find that MTKD-RL is not sensitive to RL optimization methods. Therefore, we use the original PG due to its simplicity and effectiveness.
	
	\textbf{Component analysis for RL-based optimization.}  In Table~\ref{ablation}(b), we analyse various components to construct state and action information. Using the teacher performance, including feature $\bm{f}_{i}^{T_m}$, logit $\bm{z}_{i}^{T_m}$, and cross-entropy loss $\mathcal{L}_{CE}^{T_{m}}$, leads to 1.01\% gain over AVER. Using the teacher-student gaps, including feature similarity $\cos(r(\bm{f}_{i}^{S}),\bm{f}_{i}^{T_m})$ and logit KL-divergence $\mathcal{D}_{KL}(\bm{y}_{i}^{S},\bm{y}_{i}^{T_{m}})$, results in 0.63\% gain. Gathering the comprehensive knowledge of teacher performance and teacher-student gaps for RL-based optimization achieves the best 1.56\% improvement.
	
	\textbf{Training costs.} As shown in Table~\ref{ablation}(c), we compare training complexity in one epoch with other methods over NVIDIA RTX 4090. Compared to the AVER without multi-teacher weight optimization, MTKD-RL needs extra 15\% time and 14\% memory costs, because our method requires agent training and storing (state, action, reward) as episode history. In contrast to state-of-the-art CA-MKD, MTKD-RL uses 13\% less time and extra 10\% memory footprint but achieves a better accuracy gain. In summary, our method leads to significant performance improvements with little increase in training complexity.

	\textbf{Parameter analyses of loss weights.} As shown in Fig.\ref{Gate}, we investigate the impact of logit loss weight $\alpha$ and feature loss weight $\beta$, where $\alpha=1$ and $\beta=5$ achieve the best performance.
	
	\textbf{Impact of the number of teacher networks.} Fig.\ref{Gate} shows the performance curve caused by the number of teacher networks $M$. As the teacher number $M$ increases, the performance generally improves, since more teachers could provide richer ensemble knowledge but saturates at a certain capacity of $M=4$. 
	
	\textbf{Ablation study of loss terms.} As shown in Fig.\ref{Gate}, we conduct an ablation study of loss terms in MTKD-RL. Compared to the baseline without distillation, logit-level and feature-level KD lead to 4.14\% and 3.61\% accuracy improvements, respectively. One of the reasons may be that logit-level KD is architecture-unaware distillation and more robust than feature-level KD. Summarizing the logit-level and feature-level KD provides comprehensive information for the student, maximizing the accuracy gain by 5.55\%.

	\begin{table}[t!]
		\centering
		
		\resizebox{0.85\linewidth}{!}{
			\begin{tabular}{l |c c c c}
				\toprule
				Method  & Baseline & MTKD-RL & +DIST &  +ND \\
				\midrule
				Acc@1 & 70.35 & 72.82 & 73.29&  \textbf{73.61} \\
				\bottomrule
		\end{tabular}}
		\caption{Combined with single-teacher KD methods to distill ResNet-18 on ImageNet. }
		\label{mkd_kd}
	\end{table}
	
	\textbf{Combined with single-teacher KD.} In Table~\ref{mkd_kd}, Combining MTKD-RL with advanced single-teacher KD methods of DIST~\cite{huang2022knowledge} and ND~\cite{wang2025improving} further achieves 0.47\% and 0.79\% accuracy gains, respectively.

	\section{Conclusion}
	
	This paper proposes MTKD-RL, formulating multi-teacher KD as an RL-based decision process. Compared to existing works, our method uses more comprehensive \emph{teacher performance} and  \emph{teacher-student gaps} to construct the input evidence. It reinforces the interaction between teacher and student by an agent. Experimental results on visual recognition tasks show that MTKD-RL achieves state-of-the-art performance among existing multi-teacher KD methods. We hope our work could inspire future research to explore more advanced RL strategies for multi-teacher KD.
	\appendix
		\section{A.1 Methodology}
	\subsection{Action Construction}
	For easy implementation, we integrate multiple agents into a single agent for action construction. The agent has a generator to optimize multi-teacher distillation weights.  The generator sequentially includes a linear layer,  a ReLU activation function and two separated linear layers finished by the softmax function to output the feature-level and logit-level multi-teacher distillation weight vectors $\bm{w}_{f}^{gen}\in \mathbb{R}^{M}$ and $\bm{w}_{l}^{gen}\in \mathbb{R}^{M}$, respectively.  We also combine the confidence- and divergence-aware weight generation strategies in the action space. The confidence-aware weight generation strategy is inspired by CA-MKD~\cite{zhang2022confidence}, where the $m$-th teacher distillation weight $w_{m}^{conf}$ is formulated as:
	\begin{equation}
		w_{m}^{conf}=\frac{1}{K-1}(1-\frac{\exp(\mathcal{L}_{m}^{CE})}{\sum_{j=1}^{M}\exp(\mathcal{L}_{j}^{CE})}).
	\end{equation}
	The confidence-aware multi-teacher distillation weight distribution is formulated as:
	\begin{equation}
		\bm{w}^{conf}=[w_{1}^{conf},w_{2}^{conf},\cdots,w_{M}^{conf}]\in \mathbb{R}^{M}
	\end{equation}
	For the divergence-aware weight generation strategy, we adopt teacher-student feature similarity (formulated as Equ.(\ref{feat_sim})) and probability KL-divergence (formulated as Equ.(\ref{t_s_logit_div})) with softmax normalization to compute multi-teacher weights. Specifically, the feature-level multi-teacher weight distribution $\bm{w}_{f}^{div}\in \mathbb{R}^{M}$ derived from the teacher-student feature similarity is formulated as:
	\begin{equation}
		\bm{w}_{f}^{div}=softmax([\cos^{T_1},\cos^{T_2},\cdots,\cos^{T_M}]).
	\end{equation}
	Here, we assign a larger weight to the teacher who has a larger feature similarity  with the student. This is because the more similar teacher-student pair means a smaller semantic gap and the teacher could provide matched semantic information to the student. The logit-level multi-teacher weight distribution $\bm{w}_{l}^{div}\in \mathbb{R}^{M}$ derived from the probability KL-divergence is formulated as:
	\begin{equation}
		\bm{w}_{l}^{div}=softmax([KL^{T_1},KL^{T_2},\cdots,KL^{T_M}]).
	\end{equation}
	Here, we assign a larger weight to that teacher who has a larger probability KL-divergence. All teachers in the pool are often more superior than the student, and produce higher-quality class probability distributions. Therefore, we guide the student to learn all teachers' final outputs, and emphasize the distillation strength to that teacher who is not well-aligned with the student.
	
	The feature-level multi-teacher distillation weight distribution $\bm{w}_{f}$ is formulated as a weighted fusion from three types of weight vectors:
	\begin{equation}
		\bm{w}_{f}=\gamma_{f}^{gen}*\bm{w}_{f}^{gen}+\gamma_{f}^{conf}*\bm{w}^{conf}+\gamma_{f}^{div}*\bm{w}_{f}^{div}.
	\end{equation}
	The logit-level multi-teacher distillation weight distribution $\bm{w}_{l}$ is formulated as a weighted fusion from three types of weight vectors:
	\begin{equation}
		\bm{w}_{l}=\gamma_{l}^{gen}*\bm{w}_{l}^{gen}+\gamma_{l}^{conf}*\bm{w}^{conf}+\gamma_{l}^{div}*\bm{w}_{l}^{div}.
	\end{equation}
	Here, $\gamma_{f}^{gen}$, $\gamma_{f}^{conf}$, $\gamma_{f}^{div}$, $\gamma_{l}^{gen}$, $\gamma_{l}^{conf}$. $\gamma_{l}^{div}$ are balancing parameters, which can be constant  or learnable. In practice, we found that simply using equal balancing parameters, \emph{i.e.}, $\gamma_{f}^{gen}=\gamma_{f}^{conf}= \gamma_{f}^{div}=\frac{1}{3}$, $\gamma_{l}^{gen}=\gamma_{l}^{conf}=\gamma_{l}^{div}=\frac{1}{3}$, works well. More detailed tuning of balancing parameters may achieve better performance. Moreover, we can also regard them as learnable parameters, and we found the overall performance improvements are similar to the constant parameters. More sophisticated algorithms to optimize learnable parameters may further facilitate the multi-teacher distillation performance.
	
	\section{A.2 Training Details}
	\subsection{Image Classification}  
	\begin{itemize}
		\item \textbf{CIFAR-100.} We adopt the standard image pre-processing pipeline~\cite{he2016deep}, \emph{i.e.} random cropping and flipping. The resolution of each input image is 32$\times$32 after pre-processing. 
		We apply Stochastic Gradient Descent (SGD) optimizer to train the network, where the momentum is 0.9, the weight decay is $1\times 10^{-4}$, the batch size is 64, and the initial learning rate is 0.05. The learning rate is multiplied by 10 after the 100-th and 150-th epoch within the total 240 epochs. Training is conducted on a single NVIDIA 4090 GPU. Experiments are implemented by Pytorch framework~\cite{paszke2019pytorch}.

		\item \textbf{ImageNet.} We adopt the standard image pre-processing pipeline~\cite{he2016deep}, \emph{i.e.} random cropping and flipping. The resolution of each input image is 224$\times$224 after pre-processing. For CNN and ViT, we conduct different training setups:
		
		(1) \textbf{CNN.} We apply SGD optimizer to train the network, where the momentum is 0.9, the weight decay is $1\times 10^{-4}$, the batch size is 256, and the initial learning rate is 0.1. The learning rate is multiplied by 10 after the 30-th, 60-th ,and 90-th epoch within the total 100 epochs.
		
		(2) \textbf{ViT.} We apply AdamW optimizer to train the network, where the weight decay is 0.05, the batch size is 1024, and the initial learning rate is 0.001. The detailed training setup follows DeiT~\cite{touvron2021training}.
		
		Training is conducted on 8 NVIDIA A800 GPUs. Experiments are implemented by Pytorch framework~\cite{paszke2019pytorch}.
		
	\end{itemize}
	\subsection{Object Detection}  
	\textbf{COCO-2017.}  We adopt the default data pre-processing of MMDetection~\cite{chen2019mmdetection}. The shorter side of the input image is resized to 800 pixels, the longer side is limited to 1333 pixels. We adopt a 2x training schedule with 24 epochs. Training is conducted on 8 NVIDIA A800 GPUs using synchronized SGD with a batch size of 1 per GPU.  
	
	\subsection{Semantic Segmentation}
	\textbf{Cityscapes.} We adopt the standard image pre-processing pipeline~\cite{yang2022cross}, \emph{i.e.}, random flipping and scaling in the range of [0.5, 2]. We apply SGD optimizer to train the segmentation network, where the momentum is 0.9, the batch size is 8, and the initial learning rate is 0.1. The learning rate is decayed by $(1-\frac{iter}{total\_iter})^{0.9}$ following the polynomial annealing policy~\cite{chen2017rethinking} within the total 80K training iterations. Training is conducted on 8 NVIDIA A800 GPUs using synchronized SGD with a batch size of 1 per GPU. Experiments are implemented by Pytorch framework~\cite{paszke2019pytorch}.

	\section{Acknowledgments}
	 This work is partially supported by the National Natural Science Foundation of China (No.62476264 and No.62406312), China National Postdoctoral Program for Innovative Talents (No.BX20240385) funded by China Postdoctoral Science Foundation, Beijing Natural Science Foundation (No.4244098), and Science Foundation of the Chinese Academy of Sciences.

	\bibliography{aaai25}
		
\end{document}